\documentclass{article}

\usepackage[preprint]{corl_2026} 
\usepackage{amsmath} 
\usepackage{graphicx} 
\usepackage{wrapfig} 
\usepackage{todonotes} 
\usepackage{enumitem} 
\usepackage[subtle]{savetrees}
\newcommand{\abbv}{{\rmfamily\textsc{CUE}}}
\title{Closing the Dynamics Gap \\ through Self-Directed Robotic Interaction}

\author{%
  {\small Xilun Zhang$^{1,*}$ \enspace Liam Achenbach$^{1,2,4,*}$ \enspace Luis Pabon$^{1,\dagger}$ \enspace Matthew Foutter$^{1,\dagger}$ \enspace Marco Pavone$^{1,3}$}\\[0.5em]
  {\small $^1$Stanford University \quad $^2$ETH Zurich \quad $^3$NVIDIA Research \quad $^4$microagi}\\[0.35em]
  {\small $^*$Project Co-Lead \qquad $^\dagger$Equal Contribution}
}

\begin{document}
\maketitle


\begin{abstract}
Agentic Real2Sim2Real systems can reconstruct a scene in simulation from camera observations, but contact-rich dynamics such as friction and restitution cannot be observed without interaction. We call this the \textit{dynamics gap} and present \abbv{}, an agentic system that closes it by interacting with the environment itself. Given only a high-level instruction and no resets, the agent builds its own perception, identifies physical parameters, and targets interactions where its simulation is least accurate. On billiards, \abbv{} reduces held-out ball-position error by 72\% relative to the strongest baseline across 25 hidden-physics conditions in simulation, and we further deploy it on a real pool table. More details can be found on the project website: \url{https://cue-dynamics.github.io/}
\end{abstract}



\section{Introduction \& Related Work}\label{sec:intro}

The advent of scalable, high-fidelity simulation~\cite{gao2026isaacsim, nvidia2025isaaclab} and Multi-Modal Large Language Models (MLLMs) endowed with emerging 3D spatial understanding~\cite{openai2026gpt6astra} have enabled a \textbf{Real2Sim2Real} operational paradigm~\cite{li2026agenticrsr, ren2026real2gym, wang2026rpg}: an autonomous agent reconstructs a digital twin of the scene in simulation, iteratively refines a policy in this digital environment, and closes the loop by executing this policy on the physical robot~\cite{huh2026homebody, xiao2026vptwin}. However, given the \textit{dynamics gap}, skills learned in these simulations are not directly transferrable to the real world.

We study the \textit{dynamics gap} in the game of billiards, a setting where table reconstruction is simple but contact-driven transitions are difficult to model apriori. Prior efforts to recover system dynamics learn from offline data ~\cite{kundu2026physimcode, chen2026robosimstudio}, inheriting human embodiment and action priors, and importantly, are passive: the reconstruction agent cannot actively choose which interactions to examine. System identification methods~\cite{huang2023compass, zhang2024dynamicsprompts}, including active exploration~\cite{memmel2024asid}, let the agent learn from interaction, but within a simulator and parameter space specified by a human, whose inductive biases can constrain the search space and limit both generalization and scalability.

We present \abbv{}, Calibration Using Exploration, an agentic Real2Sim2Real system that closes the dynamics gap in simulation by actively probing the environment, predicting object interactions and iteratively refines physical parameters in simulation to match autonomously collected rollouts. We showcase this framework by demonstrating~\abbv{} learns the unknown pool dynamics in the real world with zero human involvement. 
\abbv{} aims to unify the action and world understanding into one system through iterative interactions in both simulation and the real world.
Our contributions are summarized as follows: we show that our agent's simulation becomes increasingly predictive of real outcomes through self play, reducing held-out ball-position error by 72\% relative to the strongest static Real2Sim2Real baseline across 25 hidden-physics conditions. We also share lessons learned from deploying a highly dynamic robotic task in the real world, in the hope of inspiring future agentic systems that can operate outside strict human supervision.


\section{Methodology \& Experiments}\label{sec:method}

\abbv{} starts with a general instruction: \textit{Reconstruct the scene in simulation, then interact with the real table until confident that the simulation explains it. There would be no human reset during your exploration, so make sure to plan ahead.} This instruction forces the agent to design the rollouts to collect and system identification protocol. During the exploration phase, we let the agent run without supervision so as to demonstrate the general application of our framework, and only intervene when a potential safety hazard occurs. 


\textbf{Reconstruct and form a prior.}
From initial images, the agent builds a MuJoCo model of the table, cushions, pockets, cue, and balls. Drawing on its own knowledge and web search, it decides which dynamics parameters are likely to matter, sets initial ranges for them, and proposes a first set of generic probing shots. This plays the role of a prior: plausible, but not yet grounded in this table.

\textbf{What the agent discovered.}
From its reasoning traces, the agent built a general sim-to-real pipeline on top of the reconstruction.
To compare sim and real, it first built a camera-based ball tracker that turns each real shot into a trajectory comparable to a simulated rollout.
It then fit the simulation parameters by minimizing sim-to-real trajectory difference with CMA-ES over parallel MuJoCo instances replaying recorded shots from their observed initial states.
Instead of continuing its generic opening shots, it proposed new shots targeting the parts of the scene its fitted model explained worst.


\begin{wrapfigure}{r}{0.5\textwidth}
  \centering
  \vspace{-1.2em}
  \includegraphics[width=\linewidth]{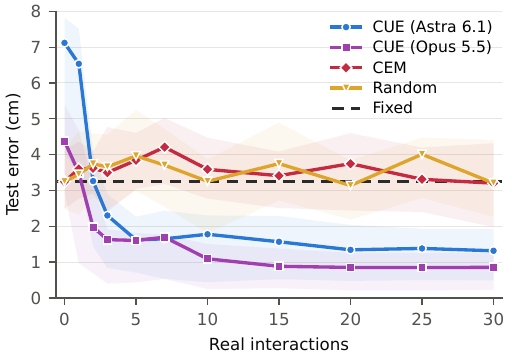}
  \vspace{-1.5em}
  \caption{Held-out ball-position error after $k$ real interactions, mean over 25 hidden-physics conditions (shaded: 25th--75th percentile).}
  \label{fig:pool_results}
  \vspace{-1em}
\end{wrapfigure}
\paragraph{Experiments}
We evaluate \abbv{} on both sim-to-sim and sim-to-real reconstruction and deployment, which demonstrate the system’s statistical performance and its practical use, respectively. 
In sim-to-sim, a MuJoCo pool simulator with hidden physics serves as the real world across 25 conditions (5 layouts $\times$ 5 physics settings), and each method gets at most 30 shots to build its model.
\abbv{} starts from an initial assumption about the dynamics, while the baselines start from the default simulation parameters (Fig.~\ref{fig:pool_results}).
After 30 shots, \abbv{} reaches a held-out ball-position error of 0.91\,cm with Opus~5.5 and 1.32\,cm with Astra~6.1, reductions of 72\% and 59\% relative to the strongest baseline (CEM, 3.20\,cm; random search, 3.20\,cm; unfitted model, 3.25\,cm).
In sim-to-real, we use a small-size pool table to validate the idea with unknown physical parameters, to further demonstrate both it's scalability and limitations in the real-world deployment. We use \abbv{} (Opus5.5) in the real-world experiment, and results are shown in the project page.

\paragraph{How to actually solve Real2Sim2Real}
\begin{itemize}[nosep, leftmargin=*]
    \item \textbf{Agentic reconstruction needs richer dynamics models.}
    Small ball imperfections and local surface variations remain difficult to reconstruct from camera views. Parameter tuning cannot capture effects outside the model, and explicitly modeling every imperfection means continually extending the simulator. A future research direction is a combination of physics-based simulation and data-driven dynamics models in agentic robotic systems.

    \item \textbf{Agentic robotics needs better experimental judgment.}
    While many claim computer vision is solved, agents still struggle to build reliable vision pipelines for real robotic experiments. Agents kept patching the vision pipeline and needed prompting to reconsider the approach. Agentic robotics creates a concrete opportunity for the vision community: robust, reusable tools that agents can reliably assemble, verify, and use in the physical world.

    \item \textbf{Agentic robotics needs stronger vision tools.}
    Our coding agent needed guidance on how to spend time and compute across long physical experiments. We need systems that preserve future exploration options and help agents decide when to think, act, refit, or backtrack in the physical world.
\end{itemize}

\clearpage
\section*{Acknowledgments}
We gratefully acknowledge support from the Defense Advanced Research Projects Agency (DARPA), the Office of Naval Research (ONR), Redwire, and microagi. We also thank the members of the Stanford Autonomous Systems Lab for their insightful discussions and valuable feedback.


\bibliography{example}  

\end{document}